\documentclass[runningheads,a4paper]{llncs}

\usepackage{amssymb}
\usepackage{amsmath}
\usepackage{graphicx}
\usepackage{color}
\usepackage{booktabs}
\usepackage{subfig}
\usepackage{url}
\usepackage{wrapfig}
\usepackage{tabularx}

\begin{document}

\mainmatter

\title{Generating Highly Realistic Images of Skin Lesions with GANs}
\titlerunning{GANs for Skin Lesion Synthesis}

\author{Christoph Baur\inst{1} \and Shadi Albarqouni\inst{1} \and Nassir Navab\inst{1,2}}
\authorrunning{Baur et al.}
\institute{Computer Aided Medical Procedures (CAMP), TU Munich, Germany\\
\and Whiting School of Engineering, Johns Hopkins University, Baltimore, United States}
\maketitle

\begin{abstract}
As many other machine learning driven medical image analysis tasks, skin image analysis suffers from a chronic lack of labeled data and skewed class distributions, which poses problems for the training of robust and well-generalizing models. The ability to synthesize realistic looking images of skin lesions could act as a reliever for the aforementioned problems. Generative Adversarial Networks (GANs) have been successfully used to synthesize realistically looking medical images, however limited to low resolution, whereas machine learning models for challenging tasks such as skin lesion segmentation or classification benefit from much higher resolution data. In this work, we successfully synthesize realistically looking images of skin lesions with GANs at such high resolution. Therefore, we utilize the concept of progressive growing, which we both quantitatively and qualitatively compare to other GAN architectures such as the DCGAN and the LAPGAN. Our results show that with the help of progressive growing, we can synthesize highly realistic dermoscopic images of skin lesions that even expert dermatologists find hard to distinguish from real ones.
\end{abstract}

\section{Introduction}

\begin{figure}
	\centering
\begin{tabular}{cc}
	\includegraphics[width=0.48\textwidth]{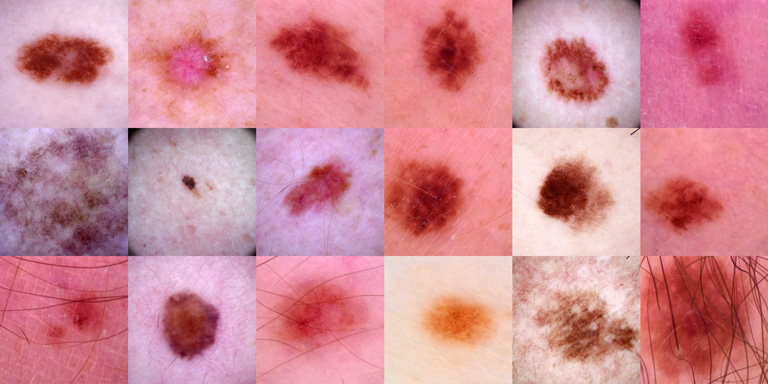} &   
	\includegraphics[width=0.48\textwidth]{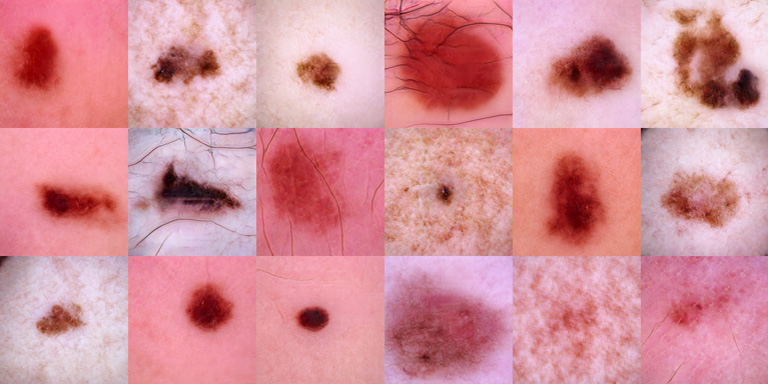} \\
(a) Real Images & (b) PGAN Samples \\[6pt]
	\includegraphics[width=0.48\textwidth]{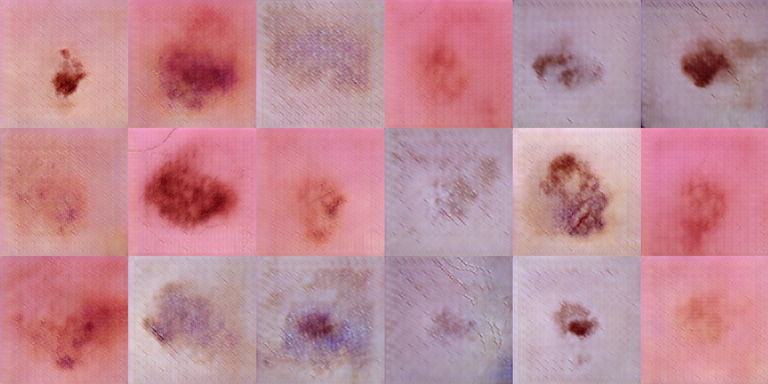} &   	\includegraphics[width=0.48\textwidth]{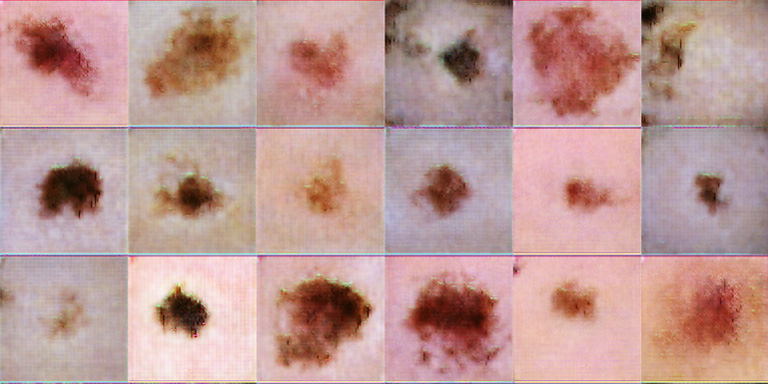} \\
(c) DCGAN Samples & (d) LAPGAN Samples \\[6pt]
\end{tabular}
	\caption{Samples generated with the different models.}
	\label{table:syntheticsamples}
\end{figure}

Just like for many other medical fields, the problems of data scarcity and class imbalance are also apparent for machine learning driven skin image analysis. In the ISIC2018 challenge, the provided dataset comprises only 10,000 labeled training samples, and the class distribution is heavily skewed among the seven categories of skin lesions, due to the rare nature of some pathologies. In order to tackle the problem of limited training data, state-of-the-art approaches for skin lesion classification and segmentation rely on heavy data augmentation~\cite{yu2017automated,matsunaga2017image} or webly supervised learning~\cite{navarro2018webly}. As an alternative, synthetic images could open up new ways to deal with these problems. Generative Adversarial Networks (GANs)~\cite{goodfellow2014generative} have shown outstanding results for this task. In the computer vision community, GANs have been successfully used for the generation of realistically looking images of indoor and outdoor scenery~\cite{radford2015unsupervised,denton2015deep}, faces~\cite{radford2015unsupervised} or handwritten digits~\cite{goodfellow2014generative}. Some conditional variants~\cite{Mirza:2014wi} have also set the new state-of-the-art in the realms of super-resolution~\cite{ledig2017photo} and image-to-image translation~\cite{Isola:2016tp}. A few of these successes have been translated to the medical domain, with applications for cross-modality image synthesis~\cite{Wolterink:2017td}, CT image denoising~\cite{Yang:2017to} and for the pure synthesis of biological images~\cite{Osokin:2017vm}, PET images~\cite{Bi:2017wf}, and OCT patches~\cite{Schlegl:2017uq}. First successful attempts for medical data augmentation using GANs have been made in \cite{antoniou2017data,frid2018synthetic}, however at a level of small patches. 

In contrast to many other medical classification problems, skin lesion segmentation and classification models often utilize ImageNet-pretrained models, meaning that these rely on input data with resolutions of $224\times 224$px or higher. For image synthesis, this implies that higher resolution images need to be generated whithout trading off realism. Thoroughly engineered, unconditional architectures such as DCGAN~\cite{radford2015unsupervised} or LAPGAN~\cite{denton2015deep} have proven to work well for high quality image synthesis from noise, however at fairly low resolution. Conditional approaches~\cite{wang2017high} have shown that both high quality and high resolution image synthesis up to $2048\times 1024$px is possible when mapping from semantic labelmaps to synthetic images with a hierarchy of conditional GANs, however this setting requires well structured input into the generator. Recently, progressive growing of GANs (PGAN)~\cite{karras2017progressive} has shown outstanding results for realistic image synthesis of faces at resolutions up to $1024\times 1024$px, without the need for any conditioning. 

\textbf{Contribution} In this work, we synthesize skin lesion images at sufficiently high resolution while ensuring high quality and realism. For our experiments, we utilize dermoscopic images of benign and malignant skin lesions provided by the ISIC2018 challenge\footnote{https://challenge2018.isic-archive.com/}. For data synthesis, we employ the PGAN and compare it to the DCGAN and the LAPGAN. As PGANs can natively only synthesize images whose size is a power of 2, we aim for a target resolution of $256\times 256$px, such that State-of-the-Art classifiers could potentially leverage the samples. A quantitative comparison of the image statistics of the synthetic and real images shows that the PGAN matches the training dataset distribution very well, and visual exploration further corroborates its superiority over the other approaches in terms of sample diversity, sharpness and artifacts. Ultimately, we evaluate the quality of the PGAN samples in a user study involving 3 expert dermatologists as well 5 Deep Learning experts, showing that the experts have a hard time distinguishing between real and fake images.

The remainder of this manuscript is organized as follows: We first briefly recapitulate the GAN framework as well as the different GAN concepts before we describe the experimental setup. Afterwards, we introduce the dataset, evaluation metrics, provide a quantitative comparison of the aforementioned concepts for skin lesion synthesis and the results of our user study. We conclude this paper with a discussion and an outlook on future work.

\section{Skin Lesion Synthesis}
\label{sec:background}

\subsection{Generative Adversarial Networks}
\label{sub:gans}

The original GAN framework consists of a pair of adversarial networks: A generator network G tries to transform random noise $z \sim p_z$ from a prior distribution $p_z$ (usually a standard normal distribution) to realistically looking images $G(z) \sim p_{fake}$. At the same time, a discriminator network D aims to classify well between samples coming from the real training data distribution $x \sim p_{real}$ and fake samples $G(z)$ generated by the generator. By utilizing the feedback of the discriminator, the generator G can be adjusted such that its samples are more likely to fool the discriminator in its classfication task, ultimately teaching the generator to approximate the training dataset distribution. Mathematically speaking, the networks play a two-player minimax game against each other:

\begin{equation}
	\min_{G} \max_{D} V(D,G) = \mathbb{E}_{x \sim p_{real}(x)}[log(D(x))] + \mathbb{E}_{z \sim p_z(z)}[1-log(D(G(z)))]
\end{equation}

In consequence, as D and G are updated in an alternating fashion, the discriminator D becomes better in distinguishing between real and fake samples while the generator G learns to produce even more realistic samples.

\begin{figure}[t]
\centering
\includegraphics[width=0.7\textwidth]{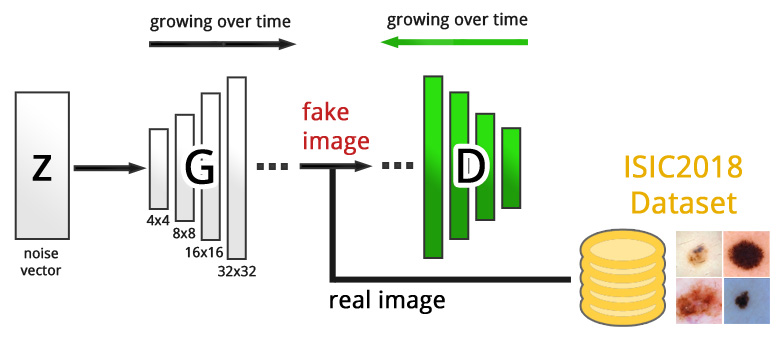}
\caption{An overview of the PGAN employed for skin lesion synthesis.}
\label{fig:pgan}
\end{figure}

In this work, we employ three different GAN concepts for the task of high resolution skin lesion synthesis, namely the DCGAN, the LAPGAN and finally the very recent PGAN. An overview of the setup is given in Fig. \ref{fig:pgan}.

\paragraph{The DCGAN}

architecture is a popular and well engineered convolutional GAN that is fairly stable to train and has proven to yield high quality results at a resolution of 64x64px. The architecture is carefully designed with concepts such as leaky ReLu activations to avoid sparse gradients and a specific weight initialization to allow for a robust training.

\paragraph{The LAPGAN}

is a generative image synthesis framework inspired by the concept of Laplacian pyramids. In essence, it consists of a hierarchy of GANs, where the first generator $G_0$ is trained to synthesize low-resolution images from noise. Successive generators $G_i$ are targeted to map from lower-resolution images of the previous generator $G_{i-1}$ to residual images, which have to be added to the upsampled, input in order to obtain compelling higher resolution images.

\paragraph{The PGAN}

utilizes the idea of progressive growing~\cite{karras2017progressive} to facilitate high resolution image synthesis from noise at unprecedented levels of quality and realism. Opposed to the LAPGAN, the PGAN consists only of a single generator and a discriminator, which both start as small networks which grow in depth and model complexity during training (see Fig. \ref{fig:pgan}). Gradually, the output-resolution of the generator and the input-resolution to the discriminator are simultaneously ramped up, leading to a very stable training behavior and very realistic, synthetic images at resolutions up to $1024\times 1024$px.

\section{Experiments and Results}
\label{sec:experiments}

In the first part of our experiments, we train a PGAN, and to prove its superiority over other concepts, also a DCGAN and a LAPGAN for skin lesion synthesis at a resolution of $256\times 256$px. In succession, we investigate the properties of the synthetic samples both quantitatively and qualitatively. In the second part of our experiments, we conduct a user study to verify the realism of the generated images.

\subsection{Dataset}

For our experiments, we utilize the ISIC2018 dataset consisting of 10,000 dermoscopic images of both benign and malignant skin lesions (see Fig. \ref{table:syntheticsamples}a). The megapixel dermoscopic images are center cropped to square size and downsampled to $256\times 256$px. No data augmentation or pre-processing was applied.

\subsection{Evaluation Metrics}
\label{sub:metrics}

A variety of methods have been proposed for evaluating the performance of GANs in capturing data distributions and for judging the quality of synthesized images. In order to evaluate visual fidelity, numerous works utilized either crowdsourcing or expert user studies. We also conduct such a user study to rate the realism of our synthetic images. In addition, we discuss visual fidelity of the generated images with a focus on diversity, realism, sharpness and artifacts. For quantitatively judging sample realism, the Sliced Wasserstein Distance (SWD) has recently shown to be a reasonably good metric for approximately comparing image distributions~\cite{karras2017progressive}, thus we also make use of it.

\subsection{Image Synthesis}
\label{sub:synthesis}

We trained a PGAN as described in \cite{karras2017progressive} from all 10,000 images, as well as a DCGAN and a LAPGAN. For all the models, the dimensionality of the latent space $\mathbf{z}$ was set to 128. The PGAN has been trained for 3M iterations, until the SWD between the synthetic samples and the training dataset did not decrease noticeably any further. For a valid comparison, the LAPGAN and DCGAN were also trained for the same amount of iterations.
For both DCGAN and LAPGAN we trained in minibatches of 8 due to GPU memory constraints on our nVidia 1080Ti, whereas for PGAN we followed the adaptive training scheme with varying minibatch sizes as proposed in the paper.

\begin{figure}[t]
\centering
\includegraphics[width=1.0\textwidth]{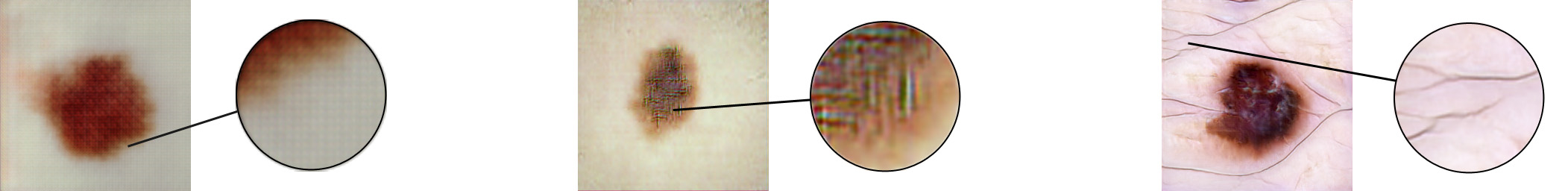}
\caption{Artifacts produced by the different models. DCGAN samples show characteristic checkerboard patterns (left), LAPGAN produces high frequency artifacts (middle), whereas PGAN has only problems synthesizing hair (right).}
\label{fig:artifacts}
\end{figure}

\begin{figure}[t]
\centering
\includegraphics[width=1.0\textwidth]{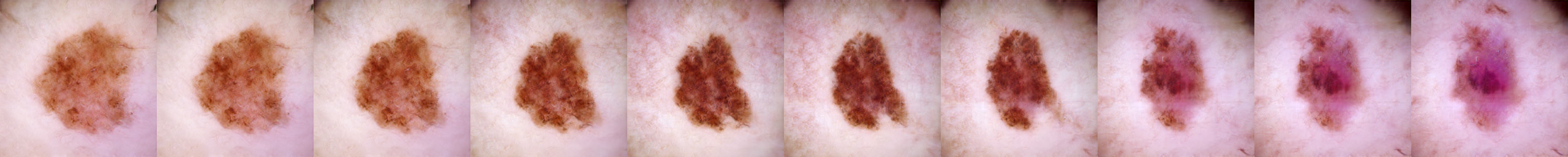}
\caption{Walking along the visual manifold of synthetic PGAN samples.}
\label{fig:manifold}
\end{figure}

Per model, we then generate 10,000 synthetic images and compare their distribution to the real data by means of the SWD (see Table \ref{table:swd}). Since the SWD constitutes an approximation, we also compute the SWD between the real data and itself to obtain a lower bound. In comparison, the lowest SWD is clearly obtained with the PGAN samples, whereas the DCGAN and LAPGAN perform considerably, but equally worse. This is also reflected by a visual exploration of the samples (see Fig. \ref{table:syntheticsamples} for a comparison of samples generated with the different models). The DCGAN samples are prone to checkerboard artifacts (Fig. \ref{fig:artifacts}, left) and can thus easily be identified as fake. The LAPGAN samples (Fig. \ref{fig:artifacts}, middle) seem more realistic and diverse, but close inspection shows a vast amount of high frequency artifacts, which again, negatively impact realism of these samples. The PGAN samples (Fig. \ref{fig:artifacts}, right) seem highly realistic, alone filamentary structures such as hair raise suspicion. 

\begin{table}[t]
	\caption{Sliced Wasserstein Distances (SWDs) between the real and generated samples from different models. Closest to the lower bound (i.e. SWD between real images and themselves) is the PGAN, whereas the distribution of DCGAN and LAPGAN samples differs considerably from the real one.}
	\label{table:swd}
	\centering
	\begin{tabularx}{\textwidth}{X|X|X|X}
		\toprule
		Lower bound 			& PGAN vs Real				& DCGAN vs Real		& LAPGAN vs Real	\\ \midrule
		4.3360						& \textbf{20.0197}		& 94.71508				& 96.68380				\\ \bottomrule
	\end{tabularx}
\end{table}


\begin{table}[t]
	\caption{Confusion matrix coefficients, Accuracy, TPR \& TNR per voter.}
	\label{table:vtt}
\begin{tabularx}{\textwidth}{@{}X|X|X|X|X|X|X|X|X@{}}
\toprule
         & DLE1   & DLE2   & DLE3   & DLE4   & DLE5   & ED1    & ED2    & ED3    \\ \midrule
TP       & 50     & 30     & 36     & 26     & 26     & 27     & 35     & 29     \\
FP       & 26     & 10     & 9      & 16     & 20     & 11     & 18     & 17     \\
FN       & 0      & 20     & 14     & 24     & 24     & 23     & 15     & 21     \\
TN       & 4      & 20     & 21     & 14     & 10     & 19     & 12     & 13     \\
ACC      & 0.675  & 0.625  & 0.712 	& 0.500  & 0.450  & 0.575  & 0.587  & 0.525  \\
TPR      & 1.000  & 0.600  & 0.720  & 0.520  & 0.520  & 0.540  & 0.700  & 0.580  \\
TNR      & 0.133  & 0.666  & 0.700  & 0.466  & 0.333  & 0.633  & 0.400  & 0.433  \\ \bottomrule
\end{tabularx}
\end{table}

\begin{table}[t]
	\caption{Fleiss' kappa scores for DLEs, EDs and all together on real, fake and all samples}
	\label{table:kappa}
\begin{tabularx}{\textwidth}{@{}X|X|X|X@{}}
\toprule
         					& DLEs   	& EDs   	& All     \\ \midrule
Fleiss' kappa (real samples)     & -0.0071  & 0.1676  & 0.0338   \\ \midrule
Fleiss' kappa (fake samples)     & 0.0338  & 0.0758  & 0.0586   \\ \midrule
Fleiss' kappa (all samples)     & 0.0265  & 0.1423  & 0.0579   \\ \bottomrule
\end{tabularx}
\end{table}

\paragraph{Exploring the Visual Manifold}
Since the PGAN samples look so compelling, there might be a chance that the model memorized the training dataset. Therefore, we explore the manifold of synthetic samples. The smooth transitions among samples provide clear evidence that memorization did not occur (see Fig. \ref{fig:manifold}).

\subsection{Visual Turing Test}
\label{sub:visual_turing_test}

In order to juge realism of the generated images, we conduct a so-called Visual Turing Test (VTT) involving 3 expert dermatologists (ED) and 5 deep-learning experts (DLE). Each participant is asked to classify the same random mix of generated and real images as being either real (class 1) or fake (class 0). The DLEs are familiar with common GAN artifacts and are thus expected to be skilled to identify unplausible generated images, even though they do not have experience in judging actual skin lesion images. On the other hand, the EDs are not aware of these deep-learning induced image artifacts, but instead know about the gamut of possible skin lesion phenotypes.

\begin{figure}[ht]
\centering
\subfloat[][TPR, FPR and Accuracies of all the voters, color coded by expert type.]{
	\centering
	\includegraphics[width=0.48\textwidth]{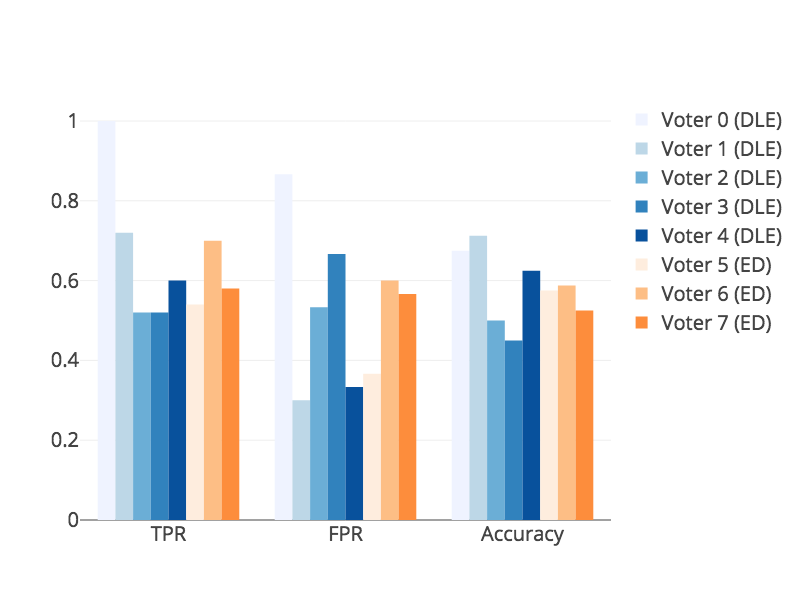}
	\label{fig:barplot}
	}
\subfloat[][Boxplots for the classification accuracies of DLEs (left) and EDs (right).]{
	\centering
	\includegraphics[width=0.48\textwidth]{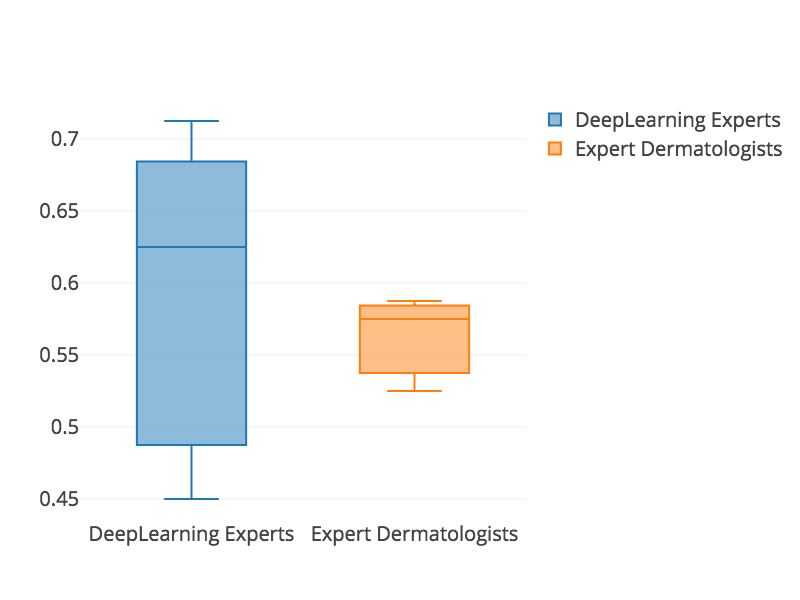}
	\label{fig:boxplot}
	}
\label{fig:test}
\caption{Visual Turing Test Results}
\end{figure}


Using the PGAN, we first generate 30 synthetic images, which are then mixed with 50 randomly chosen images from the real training dataset. In the VTT, we present each participant with these 80 images in random order and let him/her classify. Instead of training the participants up front, we give them the option to revise their classification while stepping through all the images, letting them learn and recognize patterns by themselves over time, which is also the reason for the higher number of real images. Noteworthy, in contrast to other previous works on unsupervised medical image synthesis, we do not preprocess the samples with gaussian or anisotropic filtering before presenting it to the test participants.
The performances of all the participants in terms of the TPR (how many real images have been identified as real), the FPR (how many fake images have ben classified as real) and the Accuracy are reported in Fig \ref{fig:barplot}. Performance statistics among EDs and DLEs are provided in Fig. \ref{fig:boxplot}), and the complete user study details can be found in Table \ref{table:vtt}. We also report the inter-rater agreement among DLEs, EDs and all combined by means of Fleiss' kappa~\cite{fleiss1971measuring}, for which a score of $1$ denotes perfect agreement and no agreement for values $\leq 0$. The agreement scores, reported in Table \ref{table:kappa}, show that there is higher agreement among EDs than among DLEs, but overall the score is very close to $0$, meaning there is barely any agreement among the different raters and thus no clear distinction between real and fake samples. Interestingly, the classification accuracy is slightly lower for the EDs than for the DLEs, which we amount to the fact that DLEs are more aware of GAN artifacts. Overall, the classification accuracy is just slightly above 50\%, implying that the experts can distinguish between real and fake just slightly better than chance. Thereby, not all fakes have been mistaken as real (on average 56\%), but on average 42\% of the real images have also mistakingly be identified as fake. All in all, none of the participants is able to reliably distinguish the fake samples from real ones, leading to the conclusion that these synthetic samples are in fact highly realistic.






\section{Discussion and Conclusion}
\label{sec:discussion_and_conclusion}

We have shown that with the help of PGANs, we are able to generate extremely realistic dermoscopic images, which carves open new opportunities to tackle the problems of data scarcity and class imbalance. Yet, it is unclear to which extent these synthetic data provide additional information to supervised deep learning models. In fact, a variety of questions need to be answered, such as i) whether there is an information gain in the synthetic samples over the actual training dataset, ii) if the gain is higher than using standard data augmentaton and iii) how many training images are in fact required to obtain reliable generative models. Noteworthy, we trained the PGAN ignoring the presence of different classes. For generating images along with class information, one would need to leverage labeled data and effectively train a single model per class. Further, the synthetic images are not always perfect. In particular, the methodology has to be enhanced to account for filamentary structures. In future work, we aim to perform large scale experiments and strive to answer these question.

Overall, we have shown that we can synthesize images of skin lesions at yet unprecedented levels of realism. In fact, the level of realism is so high such that experts from both the medical and the deep-learning fields were not able to reliably distinguish real images from generated ones. This leaves us confident that such synthetic data can be leveraged for new data augmentation approaches.

\bibliography{literature.short}

\begin{thebibliography}{10}
\providecommand{\url}[1]{\texttt{#1}}
\providecommand{\urlprefix}{URL }

\bibitem{antoniou2017data}
Antoniou, A., Storkey, A., Edwards, H.: Data augmentation generative
  adversarial networks. arXiv preprint arXiv:1711.04340  (2017)

\bibitem{Bi:2017wf}
Bi, L., Kim, J., Kumar, A., Feng, D., Fulham, M.: Synthesis of positron
  emission tomography (pet) images via multi-channel generative adversarial
  networks (gans). In: Molecular Imaging, Reconstruction and Analysis of Moving
  Body Organs, and Stroke Imaging and Treatment. pp. 43--51. Springer, Cham
  (2017)

\bibitem{denton2015deep}
Denton, E.L., Chintala, S., Fergus, R., et~al.: Deep generative image models
  using a laplacian pyramid of adversarial networks. In: NIPS. pp. 1486--1494
  (2015)

\bibitem{frid2018synthetic}
Frid-Adar, M., Klang, E., Amitai, M., Goldberger, J., Greenspan, H.: Synthetic
  data augmentation using gan for improved liver lesion classification. ISBI
  pp. 289--293 (2018)

\bibitem{goodfellow2014generative}
Goodfellow, I., Pouget-Abadie, J., Mirza, M., Xu, B., Warde-Farley, D., Ozair,
  S., Courville, A., Bengio, Y.: Generative adversarial nets. In: NIPS. pp.
  2672--2680 (2014)

\bibitem{Isola:2016tp}
Isola, P., Zhu, J.Y., Zhou, T., Efros, A.A.: Image-to-image translation with
  conditional adversarial networks. In: CVPR. pp. 5967--5976 (July 2017)

\bibitem{karras2017progressive}
Karras, T., Aila, T., Laine, S., Lehtinen, J.: Progressive growing of {GAN}s
  for improved quality, stability, and variation. In: ICLR (2018)

\bibitem{ledig2017photo}
Ledig, C., Theis, L., Huszar, F., Caballero, J., Cunningham, A., Acosta, A.,
  Aitken, A.P., Tejani, A., Totz, J., Wang, Z., Shi, W.: Photo-realistic single
  image super-resolution using a generative adversarial network. CVPR pp.
  105--114 (2017)

\bibitem{matsunaga2017image}
Matsunaga, K., Hamada, A., Minagawa, A., Koga, H.: Image classification of
  melanoma, nevus and seborrheic keratosis by deep neural network ensemble.
  arXiv preprint arXiv:1703.03108  (2017)

\bibitem{Mirza:2014wi}
Mirza, M., Osindero, S.: Conditional generative adversarial nets. arXiv
  preprint arXiv:1411.1784  (2014)

\bibitem{navarro2018webly}
Navarro, F., Conjeti, S., Tombari, F., Navab, N.: Webly supervised learning for
  skin lesion classification. arXiv preprint arXiv:1804.00177  (2018)

\bibitem{Osokin:2017vm}
Osokin, A., Chessel, A., Carazo-Salas, R.E., Vaggi, F.: Gans for biological
  image synthesis. ICCV pp. 2252--2261 (2017)

\bibitem{radford2015unsupervised}
Radford, A., Metz, L., Chintala, S.: Unsupervised representation learning with
  deep convolutional generative adversarial networks. CoRR  abs/1511.06434
  (2015)

\bibitem{Schlegl:2017uq}
Schlegl, T., Seeb{\"o}ck, P., Waldstein, S.M., Schmidt-Erfurth, U., Langs, G.:
  Unsupervised anomaly detection with generative adversarial networks to guide
  marker discovery. In: IPMI. pp. 146--157. Springer, Cham (2017)

\bibitem{wang2017high}
Wang, T.C., Liu, M.Y., Zhu, J.Y., Tao, A., Kautz, J., Catanzaro, B.:
  High-resolution image synthesis and semantic manipulation with conditional
  gans. In: CVPR (June 2018)

\bibitem{Wolterink:2017td}
Wolterink, J.M., Dinkla, A.M., Savenije, M.H.F., Seevinck, P.R., van~den Berg,
  C.A.T., I{\v{s}}gum, I.: Deep mr to ct synthesis using unpaired data. In:
  Simulation and Synthesis in Medical Imaging. pp. 14--23. Springer, Cham
  (2017)

\bibitem{Yang:2017to}
Yang, Q., Yan, P., Zhang, Y., Yu, H., Shi, Y., Mou, X., Kalra, M.K., Zhang, Y.,
  Sun, L., Wang, G.: Low-dose ct image denoising using a generative adversarial
  network with wasserstein distance and perceptual loss. IEEE Transactions on
  Medical Imaging  37(6),  1348--1357 (June 2018)

\bibitem{yu2017automated}
Yu, L., Chen, H., Dou, Q., Qin, J., Heng, P.A.: Automated melanoma recognition
  in dermoscopy images via very deep residual networks. IEEE transactions on
  medical imaging  36(4),  994--1004 (2017)

\end{thebibliography}
\bibliographystyle{splncs03}

\end{document}